# Architecture for Pseudo Acausal Evolvable Embedded Systems

Mohd Abubakr and Rali Manikya Vinay,
Electronics and Communication Engineering
Gokaraju Rangaraju Institute of Engineering and Technology
Miyapur, Hyderabad, 500045 INDIA
Email id: mohdabubakr@ieee.org

*Abstract-Advances in semiconductor technology are contributing to the increasing complexity in the design of embedded systems. Architectures with novel techniques such as evolvable nature and autonomous behavior have engrossed lot of attention. This paper demonstrates conceptually evolvable embedded systems can be characterized basing on acausal nature. It is noted that in acausal systems, future input needs to be known, here we make a mechanism such that the system predicts the future inputs and exhibits pseudo acausal nature. An embedded system that uses theoretical framework of acausality is proposed. Our method aims at a novel architecture that features the hardware evolability and autonomous behavior alongside pseudo acausality. Various aspects of this architecture are discussed in detail along with the limitations.*

I. INTRODUCTION

Consumer demands are main source of inspiration for innovative designs of embedded systems. With the increase in demand for the dependable machines, new architectures for autonomous devices have emerged in the market. These devices include smart phones, smart cars that have the capability of taking decision on their own. The success of these devices has led the industry to invest towards extending autonomous behavior of machine for all applications. Research is now seeking a course which makes machines much more superior and intelligent that are not only 'smart' enough to take decision but can forecast the possible robust surroundings it faces and acclimatizes itself by modifying itself at the hardware level. Evolvable embedded systems are one of such highly demanded and invested field in the present research scenario. 'Evolvable' implies the autonomous behavior of the system to be capable of repairing and formulating a solution on its own. This autonomous behavior can be achieved through genetic programming and artificial intelligence.

In this paper, a new class of embedded systems is discussed that have certain distinguishable properties. The evolvable concept of embedded systems has been studied recently and a generic methodology has been developed [1]. We discuss the possibility of using the theoretical framework of acausality in developing evolvable embedded systems and propose a novel architecture for implementing the same.

The paper is organized as follows. Motivation and prior work discussed in section II. A brief explanation about acausality in section III. A proposed architecture to implement the projected technique has been furnished in section IV. In the sub-sections each block of the architecture is briefly discussed. The working of the architecture is explained in the section V followed by conclusion.

Research funded by: GRIET, Miyapur, Hyderabad.

II. MOTIVATION AND PRIOR WORK

Conventional embedded systems consist of a microcontroller and DSP components realized using Field programmable gate arrays (FPGA), Complex programmable logic arrays (CPLDs), etc. With the increasing trend of System on Chip (SoC) integrations, mixed signal design on the single chip has become achievable. Such systems are excessively used in the areas of wireless communication, networking, signal processing, multimedia and networking. In order to increase the quality of service (QoS) the embedded system needs to be fault tolerant, must consume low power, must have high life time and should be economically feasible. These services have become a common specification for all the embedded systems and consequently to attract attention from commercial market different researchers have come up with novel solutions to redefine the QoS of embedded systems. Future embedded systems consists of evolutionary techniques that repair, evolve and adapt themselves to the conditions they are put in. Such systems can be termed as autonomous designs.

Autonomous designs using genetic algorithms and artificial intelligence evolve new hardware systems basing on the conditions have generated interest in recent days. These systems are based upon the adaptive computational machines that produce an entirely new solution on its own when the environment gets hostile. Here hostile environment refers to the changes in temperature, increase in radiation content, under these conditions an autonomous systems needs to have an ability to modify and evolve hardware that is less susceptible to the hostile environment. Classification of embedded systems as given in [1] is as

*1)* Class 0 (fixed software and hardware): Software as well as hardware together are defined at the design time. Neither reconfiguration nor adaptation is performed. This class also contains the systems with reconfigurable FPGAs that are only configured during reset. A coffee machine could be a good example.

*2)* Class 1 (reconfigurable SW/HW): Software or hardware (a configuration of an FPGA) is changed during the run in order to improve performance and the utilization of resources (e.g. in reconfigurable computing). Evolutionary algorithm can be used to schedule the sequence of configurations at the compile time, but not at the operational time.

*3)* Class 2 (evolutionary optimization): Evolutionary algorithm is a part of the system. Only some coefficients in SW (some constants) or HW (e.g. register values) are evolved, i.e. limited adaptability is available. Fitness calculation and genetic operations are performed in software. Example: an adaptive filter changing coefficients for a fixed structure of an FIR filter.

*4) Class 3a (evolution of programs):* Entire programs are constructed using genetic programming in order to ensure adaptation or high-performance computation. Everything is performed in software [2].

*5) Class 3b (evolution of hardware modules):* Entire hardware modules are evolved in order to ensure adaptation, high-performance computation, fault-tolerance or low-energy consumption. Fitness calculation and genetic operations are carried out in software or using a specialized hardware. Reconfigurable hardware is configured using evolved configurations. The system typically consists of a DSP and a reconfigurable device. Example: NASA JPL SABLES [3].

*6) Class 4 (evolvable SoC):* All components of class 3b are implemented on a single chip. It means that the SoC contains a reconfigurable device. Some of such devices have been commercialized up to now, for example, a data compression chip [4].

*7) Class 5 (evolvable IP cores):* All components of class 3b are implemented as IP cores, i.e. at the level of HDL source code (Hardware Description Language). It requires describing the reconfigurable device at the HDL level as well. An approach—called the virtual reconfigurable circuit—has been introduced to deal with this problem [15]. Then the entire evolvable subsystem can be realized in a single FPGA.

*8) Class 6 (co-evolving components):* The embedded system contains two or more co-evolving hardware or software devices. These co-evolving components could be implemented as multiprocessors on a SoC or as evolvable IP cores on an FPGA. No examples representing this class are available nowadays.

Figure 1: Evolvable component placed in evolvable embedded system. Components 1–4 represent the environment for the evolvable component in this example.

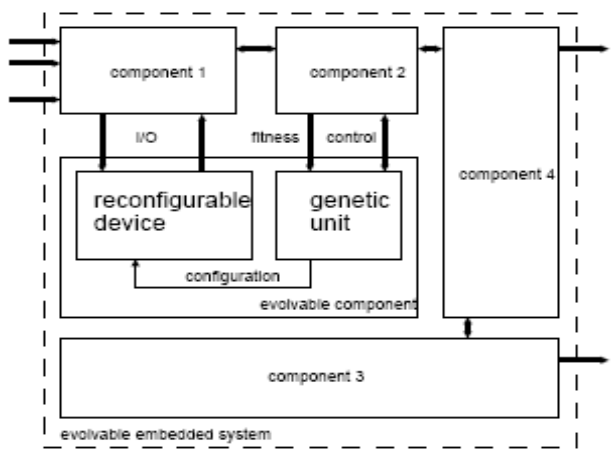

Any embedded system can be categorized through the classes given above. Reconfigurable computing has significantly contributed to the idea of have evolvable hardware through dynamically upload/remove the hardware components from the hardware module library. According to [1], an evolvable embedded system can be defined as "a reconfigurable embedded system in which an evolutionary algorithm is utilized to dynamically modify some of system (software and/or hardware) components in order to adapt the behavior of the system to a changing environment". Figure 1 shows a general block diagram for evolvable embedded system.

III. ACAUSALITY

Embedded systems design is based on the assumption that it generates an output based on present and past inputs. These set of embedded systems are called as real time embedded systems. The new set of embedded systems that is introduced here is based on the assumption that the output is generated considering even the future inputs. Such systems can be termed as Acausal Embedded Systems. Due to the uncertainty in prediction of the future inputs these systems require a specialized master algorithm that evolves hardware to implement specified output, and this evolved hardware is constructed via available resources.

We define acausality as the term denoted to the systems whose present outputs are dependent on past, present and future inputs. Acausality is in total contrast with the present convention of embedded systems that purely rely on present and past inputs.

IV. PROPOSED SYSTEM

Various proposals for autonomous embedded systems are available in the literature [16]. The system proposed here belongs to Class 6 group of systems. It can evolve hardware and software on its own using artificial intelligence algorithms and available set of resources. Another major block of the proposed architecture is the use of future input predictor. The details about the future input predictors are explained in the next sections. Use of artificial intelligence in determining suitable solution for the predicted future inputs is essential and plays an important role in determining the efficiency of the system.

The figure 2 shows the proposed architecture of an acausal system belonging to the class6 group of evolvable embedded systems. It can evolve hardware and software on its own using artificial intelligence algorithms and available set of resources. The contemporary technology allows the embedded hardware creator to make use of the reconfigurable hardware resources to build in an evolved design. Some reliable reconfigurable hardware platforms can be Programmable Analog Array, Field Programmable Gate Array (FPGA), FPAA (Field Programmable Analog Array), FPTA (Field Programmable Transistor Array), nano devices, reconfigurable antennas, MEMS (micro electromechanical systems), reconfigurable optics and some other few selective components.

The next sections give the generic description about each block present in the proposed architecture shown in the figure 2.

*A. Past Input Summarizer*

Functionality of the past input summarizer (PIS) is to store the past values of inputs into allocated memory. Since it is practically impossible to store all the past inputs in a memory, hence we use an algorithm to compress and summarize the past inputs. Hence we can define PIS as "it stores relevant information of the past inputs which can be applicable in predicting the future inputs, in a limited memory by following an optimized algorithm". The compression and summarization algorithms can be chosen based upon the sensitivity of the application. Conventional embedded systems have very limited memory due to the space constraints. Utilization of newly developed optical memory concept can be beneficial in increasing the potential of past inputs storage [17].

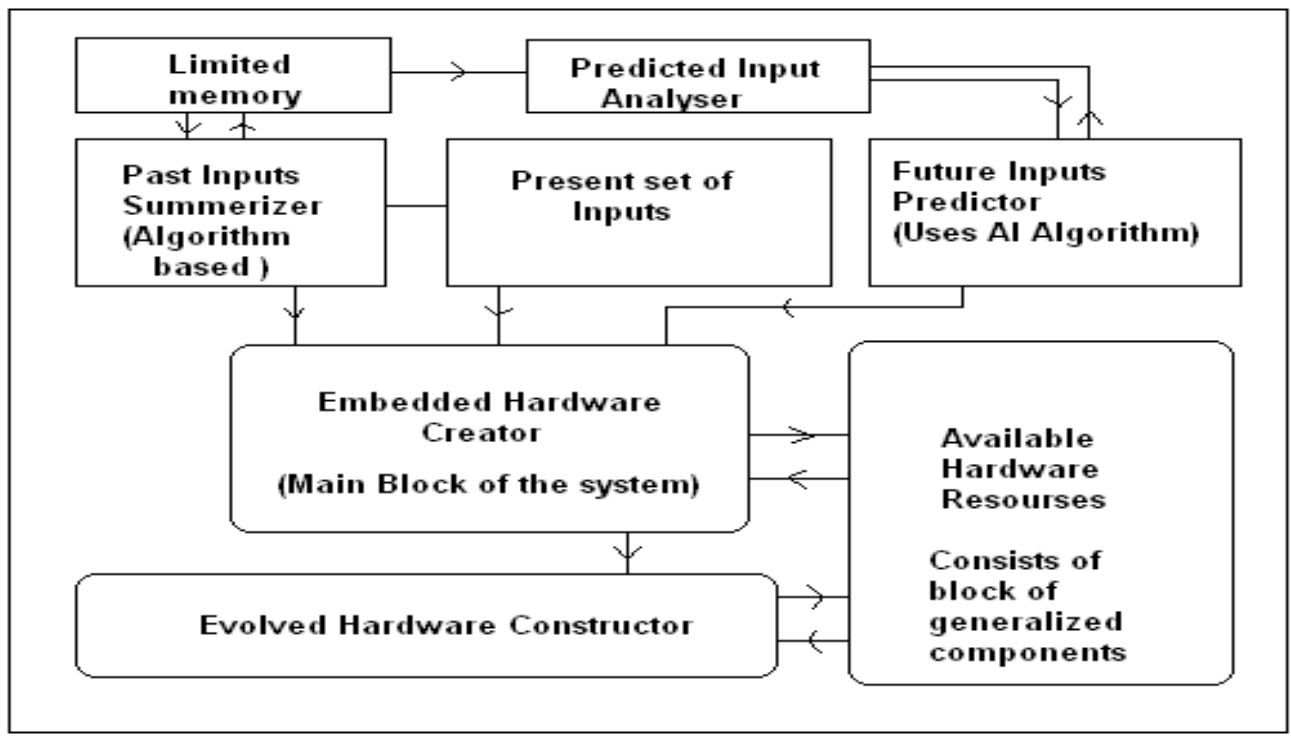

Figure 2: The generalized architecture of the proposed model

*B. Present Input*

Acquisition of present inputs is the fundamental functionality of this block. There may be wide varieties of inputs depending on the applications i.e. the inputs such as signals obtained through sensor elements, transmission receiver, transducer, etc. At this juncture noise factors are taken into consideration; consequently this block plays an important role in determining the efficiency of the intact embedded system. After execution of a particular input it is sent to past input summarizer and a new input is extracted. It is a requisite that noise gets eliminated at this block itself else the noise subsumes at past input summarizer.

*C. Future Input Predictor and Analyzer*

Future input predictor (FIP) block as the name indicates predicts the future inputs. Here an allotment is made to pass on future inputs through another device as well. This allotment is done using either antenna or some special sensor networks. Sometimes, future inputs can be known through external agents, hence this allotment forms an interface between the external agent and the system. This provision makes this block exhibit dual stability in predicting the future inputs. Many algorithms for future prediction are available in literature. Future prediction and estimates are extensively used in financial decisions. Future inputs can also be predicted using the data obtained from past inputs. Using pattern recognition algorithms for discrete sets of past inputs, the future inputs can be determined. There are plenty of algorithms that are develops to solve pattern recognition from the available data [5,6,7,8].

*D. Embedded Hardware Creator*

This is the heart of acausal embedded system, as it is the most important block of all. Embedded Hardware creator (EHC) should have an autonomous capability i.e. should be capable of making decisions. This can be achieved using advance techniques such as artificial intelligence (adaptable neural networks), genetic computing, evolutionary computing etc.

Artificial Neural networks (ANN) are models of human brain used to perform tasks based on self-learning and adaptation to environment. Creating ANNs that can learn and generalize the information from surrounding is the first step in having autonomous computational machines. Due to practical inabilities the learning time for ANNs is very high [9,10]. Another mechanism that can help in realizing EHC is cellular automata. Cellular automata are discrete spatially extended dynamical systems that have been extensively studied as a model of computational devices [11,12]. Evolutionary algorithms such as genetic algorithms have generated a huge interest in this area. Genetic algorithms are based upon selection and mutation [13,14]. Such techniques can be useful in implementing EHC.

*E. Available Hardware Resources*

For the construction of the hardware devices certain amount of resources are allocated to the embedded system creator. These allocated resources are based upon the future need for up-gradation and cost of the total embedded system. FPGA, FPAA, PALS, memory elements, etc are few such examples of reconfigurable devices that can be used as resources for the embedded hardware creator. A provision is provided to increase or repair the available resources in order to upgrade the system at any moment. Size of the available resources is a constraint since it induces the tendency of the system to be more susceptible to thermal noise effects and radiation. In order to reduce the unwanted noise effects caused by thermal variance and incoming radiation proper shielding techniques are needed else available resources may become defective. Another parameter that determines the number of available resources is the cost. The economic feasibility of the embedded system is crucial for the commercial success.

*F. Evolved Architecture constructor*

This block is the final evolved design done by the embedded hardware creator. This block has inbuilt design features to execute the instructions given by EHC into building up an actual hardware design. It also consists of read/write/erase mechanisms for FPGA, FPAA, FPTA and other memories. Hence it forms an interface between EHC and Available resources.

## V. WORKING OF THE SYSTEM

The sequence of operations that go in the system are as follows. First the past input summarizer (PIS), the present input and the future input predictor (FIP) give realizable inputs to the embedded hardware creator and then by following a relevant logic a layout for the new derived design results as an outcome from the embedded hardware creator depending upon the inputs from PIS, present input and FIP. The EHC is accountable to sending instructions to EAC about the construction of the system using available hardware resources. After receiving the instructions from EHC, EAC launches the concrete effort of writing and erasing of the reconfigurable hardware resources, connecting interconnects, etc to construct the working structure of the predicted solution. The potential capacity of the hardware depends on the type of system that is used. The construction of hardware design is not a single step process but a continuous process that repeats itself for a more resourceful and well-organized design. The proposed system is a generalized version that can be modified as per the application in which the concept of acausal self-evolving reconfigurable hardware is used.

## VI. CONCLUSION

In this paper we have proposed a model that uses the theoretical framework of acausality. The proposed architecture is a generalized version of evolvable architectures and basing on the application it can be suitably modified. The proposal of pseudo acausal evolvable embedded systems opens up a path for a new era of research and the pace of technological changes assume a new shape where we find the machines repairing themselves and evolving autonomously removing the major bottle-necks of maintenance and non-durable nature of the existing embedded systems. This implementation of such technology finds itself an imperative place in every field of application. Some if its prospective aspects viewed in near future are in aeronautics, astronautics, robotics, etc. This technology may develop as a capstone for evolvable embedded system applications and AI research.

The generalized concept of modeling evolvable embedded systems have been realized in terms of reconfigurable components and artificial intelligence, our future research will be in creating tools for such design. Due to financial constraints we have restricted our work only up to theoretical work and we hope in near future to practically demonstrate such a system.